\documentclass[pdflatex,sn-vancouver-num]{sn-jnl}

\usepackage{graphicx}
\usepackage{booktabs}
\usepackage{amsmath,amssymb}
\usepackage{multirow}
\usepackage{array}
\usepackage{tabularx}
\usepackage{makecell}      
\usepackage{indentfirst}
\usepackage{enumitem}
\usepackage{pdfpages}
\setlist[itemize]{left=\parindent}   
\setlist[enumerate]{left=\parindent} 
\newcolumntype{P}[1]{>{\raggedright\arraybackslash}p{#1}}
\newcolumntype{Y}{>{\raggedright\arraybackslash}X} 

\begin{document}

\title{Radiology’s Last Exam (RadLE): Benchmarking Frontier Multimodal AI Against Human Experts and a Taxonomy of Visual Reasoning Errors in Radiology}

\author*[1]{\fnm{Suvrankar} \sur{Datta}}\email{suvrankar.datta@ashoka.edu.in}
\equalcont{These authors contributed equally to this work.}
\author*[1]{\fnm{Divya} \sur{Buchireddygari}}\email{divyabuchireddy@gmail.com}
\equalcont{These authors contributed equally to this work.}

\author[1]{\fnm{Lakshmi Vennela Chowdary} \sur{Kaza}}
\author[1]{\fnm{Mrudula} \sur{Bhalke}}
\author[1]{\fnm{Kautik} \sur{Singh}}
\author[1]{\fnm{Ayush} \sur{Pandey}}
\author[1]{\fnm{Sonit Sai} \sur{Vasipalli}}
\author[1]{\fnm{Upasana} \sur{Karnwal}}
\author[1]{\fnm{Hakikat Bir Singh} \sur{Bhatti}}
\author[1]{\fnm{Bhavya Ratan} \sur{Maroo}}
\author[1]{\fnm{Sanjana} \sur{Hebbar}}
\author[1]{\fnm{Rahul} \sur{Joseph}}
\author[1]{\fnm{Gurkawal} \sur{Kaur}}
\author[2]{\fnm{Devyani} \sur{Singh}}
\author[1]{\fnm{Akhil} \sur{V}}
\author[1]{\fnm{Dheeksha Devasya Shama} \sur{Prasad}}
\author[1]{\fnm{Nishtha} \sur{Mahajan}}
\author[1]{\fnm{Ayinaparthi} \sur{Arisha}}
\author[2]{\fnm{Rajesh} \sur{Vanagundi}}
\author[1]{\fnm{Reet} \sur{Nandy}}
\author[1]{\fnm{Kartik} \sur{Vuthoo}}
\author[1]{\fnm{Snigdhaa} \sur{Rajvanshi}}
\author[1]{\fnm{Nikhileswar} \sur{Kondaveeti}}
\author[1]{\fnm{Suyash} \sur{Gunjal}}
\author[2]{\fnm{Rishabh} \sur{Jain}}
\author[2]{\fnm{Rajat} \sur{Jain}}
\author[3]{\fnm{Anurag} \sur{Agrawal}}

\affil[1]{\orgdiv{Centre for Responsible Autonomous Systems in Healthcare (CRASH) Lab, Koita Centre for Digital Health}, \orgname{Ashoka University}, \orgaddress{\city{Sonipat}, \country{India}}}

\affil[2]{\orgname{Independent Researcher}, \orgaddress{\city{New Delhi}, \country{India}}}

\affil[3]{\orgdiv{Koita Centre for Digital Health}, \orgname{Ashoka University}, \orgaddress{\city{Sonipat}, \country{India}}}

\abstract{
Generalist multimodal AI systems such as large language models (LLMs) and vision language models (VLMs) are increasingly accessed by clinicians and patients alike for medical image interpretation through widely available consumer-facing chatbots. Most evaluations claiming expert level performance are often on public datasets containing more common pathologies and fail to reflect the complexity of real-world radiology, which demands the detection of subtle and challenging cases. Rigorous evaluation of frontier models on difficult diagnostic cases remains limited.
We developed a pilot benchmark of 50 expert-level “spot diagnosis” cases across multiple imaging modalities to evaluate the performance of frontier AI models against board-certified radiologists and radiology trainees. To mirror real-world usage patterns, the reasoning modes of five popular frontier AI models were tested through their native web interfaces, viz. OpenAI o3, OpenAI GPT-5, Gemini 2.5 Pro, Grok-4, and Claude Opus 4.1. Accuracy was scored by blinded experts, and reproducibility was assessed across three independent runs.
GPT-5 was additionally evaluated across various reasoning modes through the API. Reasoning quality errors were assessed separately by independent raters and a consensus taxonomy of visual reasoning errors was defined. Board-certified radiologists achieved the highest diagnostic accuracy (83\%), outperforming trainees (45\%) and all AI models (best performance shown by GPT-5: 30\%). Reliability was substantial for GPT-5 and o3, moderate for Gemini 2.5 Pro and Grok-4, and poor for Claude Opus 4.1.
These findings demonstrate that advanced frontier models fall far short of radiologists in challenging diagnostic cases. Our benchmark highlights the present limitations of generalist AI in medical imaging and cautions against unsupervised clinical use. We also provide a qualitative analysis of reasoning traces and propose a practical taxonomy of visual reasoning errors by AI models as an essential step toward understanding their failure modes, informing evaluation standards and guiding more robust model development.

}


\maketitle

\begin{figure}[htbp]
    \centering
    \includegraphics[width=0.9\linewidth]{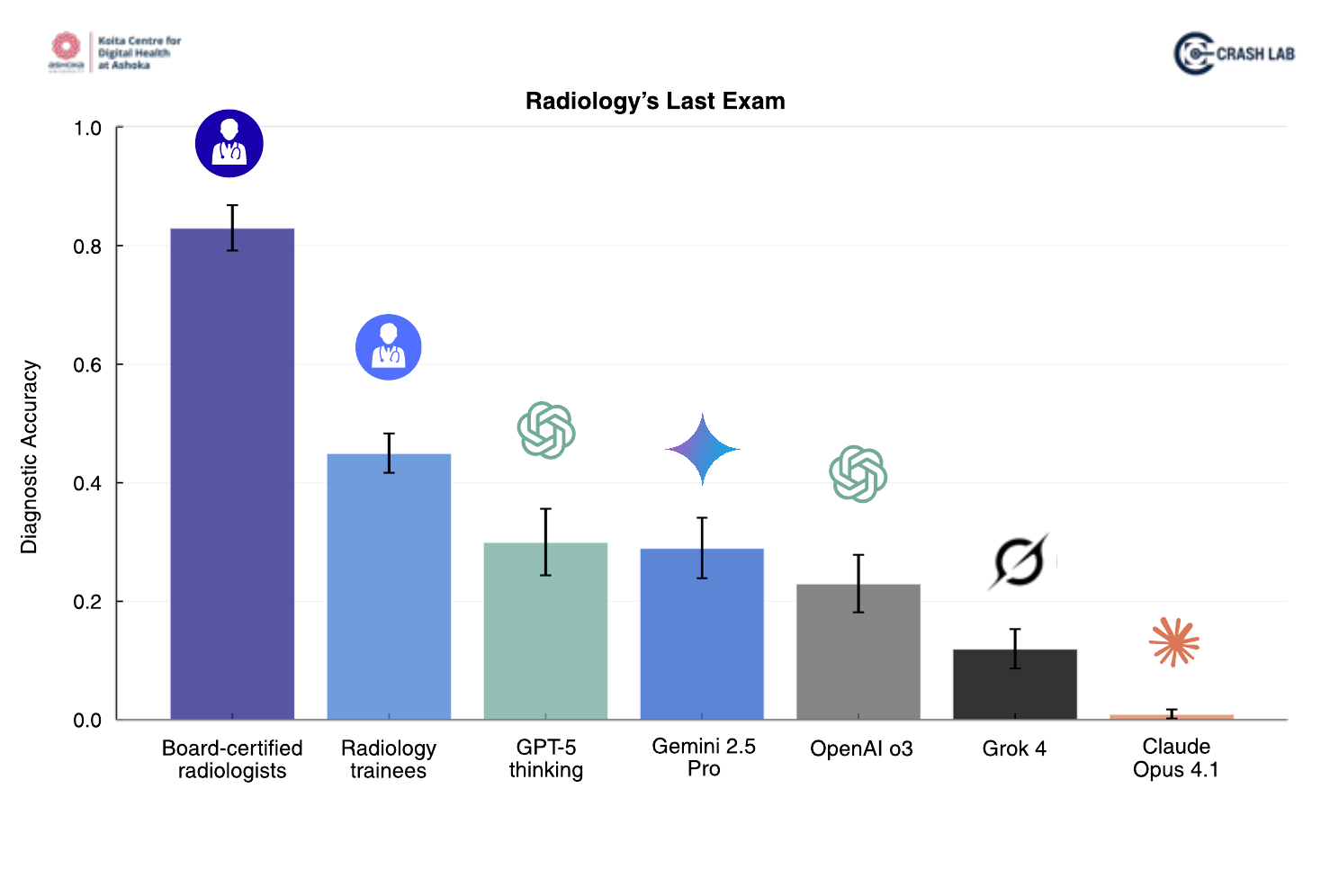}
    \caption{Diagnostic accuracy across humans and multimodal AI systems on the Radiology’s Last Exam (RadLE) v1 benchmark. Board-certified radiologists achieved the highest accuracy (0.83), followed by trainees (0.45). All tested frontier models underperformed, with GPT-5 (0.30) and Gemini 2.5 Pro (0.29) showing the best AI results but falling well below human benchmarks.}
\end{figure}

\section{Introduction}\label{sec1}

Recent advances in generalist multimodal large language models (LLMs) have brought these systems into mainstream clinical discourse, with both healthcare professionals and patients now using them for medical image interpretation\cite{tu2024towards, moor2023foundation, ye2025multimodal, zhang2025revolutionizing}. Vendor reports and recent model launches frequently highlight ``expert-level'' capabilities, but such reports are often anecdotal and typically derived from evaluations on clean, single-task datasets, such as CheXpert or MIMIC-CXR, which predominantly feature common pathologies with clear visual manifestations\cite{liu2024gpt4v_cxr, liu2023exploring}. These datasets are enriched for common pathologies and therefore underestimate the cognitive and perceptual challenges of real-world radiology, where subtle findings, ambiguous presentations and the integration of clinical context are often required to arrive at the correct diagnosis.

The public availability of frontier AI systems through consumer-facing applications has amplified their reach, extending not only to practicing radiologists but also directly to patients\cite{robertson2023attitudes, bajaj2024chatgpt}. Anecdotal reports suggest that trainees increasingly rely on these tools in daily practice, sometimes even in patient-facing contexts. Patients, in turn, are beginning to upload scans to these platforms and occasionally prioritize AI-generated interpretations over physician consultations. This emerging trend heightens the urgency of understanding not just whether these models can identify gross abnormalities, but also whether they can handle the complex, nuanced cases that routinely challenge human experts.

To address this gap, we introduce a deliberately spectrum-biased preliminary benchmark \textit{Radiology's Last Exam (RadLE) v1}, comprising complex and high-value spot diagnostic cases curated to represent the kind of studies that differentiate novice from expert performance. We compare the diagnostic performance of frontier generalist AI models against a stratified cohort of human readers, from first-year residents to senior radiologists, to determine where current AI systems fall within the human learning curve.

\subsection{Clinical Context and Motivation for our Study}

A pilot survey of 10 radiology trainees and practicing radiologists revealed widespread use of consumer AI applications for case discussions and preliminary image interpretation of difficult studies. Models from OpenAI, Gemini, Grok and Claude were frequently accessed through mobile interfaces by trainees for decision assistance, representing a shift from traditional peer consultation toward AI-assisted problem solving. In parallel, patient-driven use of these same systems has also become more common, with some patients informally substituting AI outputs for professional consultations. These shifts raise important questions about diagnostic accuracy, accountability and clinical safety\cite{cadth2025watchlist, shekar2024overtrust}.

\subsection{Evaluation Challenges}

Most prior AI evaluations in radiology have focused on public CXR datasets like CheXpert or MIMIC-CXR, and often assess pattern recognition on common pathologies rather than expert-level diagnostic reasoning across multi-modality datasets\cite{liu2024gpt4v_cxr, liu2023exploring}. Real-world radiology practice often requires identifying subtle findings and complex cases, capabilities that often extend beyond simple classification tasks. Recent developments in LLM ``reasoning'' capabilities claim to improve performance through ``thinking'' or extended deliberation. However, systematic evaluation of these reasoning models in radiological contexts remains limited.

\subsection{Study Objectives}

This study benchmarks the diagnostic performance of frontier generalist multimodal AI models against radiologists across the expertise spectrum on a curated set of challenging spot-diagnosis cases. In addition to quantitative comparisons, we systematically examine the reasoning traces produced by these models and propose a concise taxonomy of visual reasoning errors. Together, these analyses provide evidence-based insights into the current capabilities and limitations of frontier AI in complex medical image interpretation.

\section{Methods}

\subsection{Dataset curation}
The dataset of single radiological spot-diagnosis cases was assembled via a crowdsourcing initiative involving radiologists and residents from multiple institutions, with all images de-identified prior to inclusion. Two board-certified radiologists (each with more than five years of clinical experience) independently reviewed all submissions and selected 50 cases to constitute version one of the benchmark, \textit{Radiology’s Last Exam (RadLE) v1}.
Cases were included if they (i) reflected complex diagnostic scenarios of regular radiology practice or board exams and were encountered by both reviewers within the last five years, and (ii) had a single, unambiguous reference diagnosis derivable from the imaging provided, without requiring ancillary laboratory or procedural data. Cases with broad differential diagnoses or those requiring multimodality correlation for a definitive answer were excluded.

The final v1 dataset comprised 50 radiological images spanning three imaging modalities (radiography, CT, and MRI) and six major clinical systems (cardiothoracic, gastrointestinal, genitourinary, musculoskeletal, head \& neck/neuro, and paediatric). Detailed distributions are provided in Tables 1 and 2. Reverse image search confirmed no detectable duplication with publicly available datasets.

\begin{table}[ht]
\section*{Table 1. Dataset composition by imaging modality}
\centering
\begin{tabular}{p{0.35\linewidth} p{0.25\linewidth} p{0.25\linewidth}}
\toprule
\textbf{Modality} & \textbf{Number of Cases (n)} & \textbf{Percentage (\%)} \\
\midrule
Radiograph (X-ray) & 13 & 26\% \\
Computed Tomography (CT) & 24 & 48\% \\
Magnetic Resonance Imaging (MRI) & 13 & 26\% \\
\midrule
\textbf{Total} & \textbf{50} & \textbf{100\%} \\
\botrule
\end{tabular}
\end{table}

\begin{table}[ht]
\section*{Table 2. Dataset composition by clinical system}
\centering
\begin{tabular}{p{0.35\linewidth} p{0.25\linewidth} p{0.25\linewidth}}
\toprule
\textbf{Clinical System} & \textbf{Number of Cases (n)} & \textbf{Percentage (\%)} \\
\midrule
Cardiothoracic & 7 & 14\% \\
Gastrointestinal & 8 & 16\% \\
Genitourinary & 7 & 14\% \\
Musculoskeletal & 9 & 18\% \\
Head \& Neck / Neuro & 9 & 18\% \\
Paediatric & 10 & 20\% \\
\midrule
\textbf{Total} & \textbf{50} & \textbf{100\%} \\
\botrule
\end{tabular}
\end{table}

\begin{figure}[htbp]
    \centering
    \includegraphics[width=0.9\linewidth]{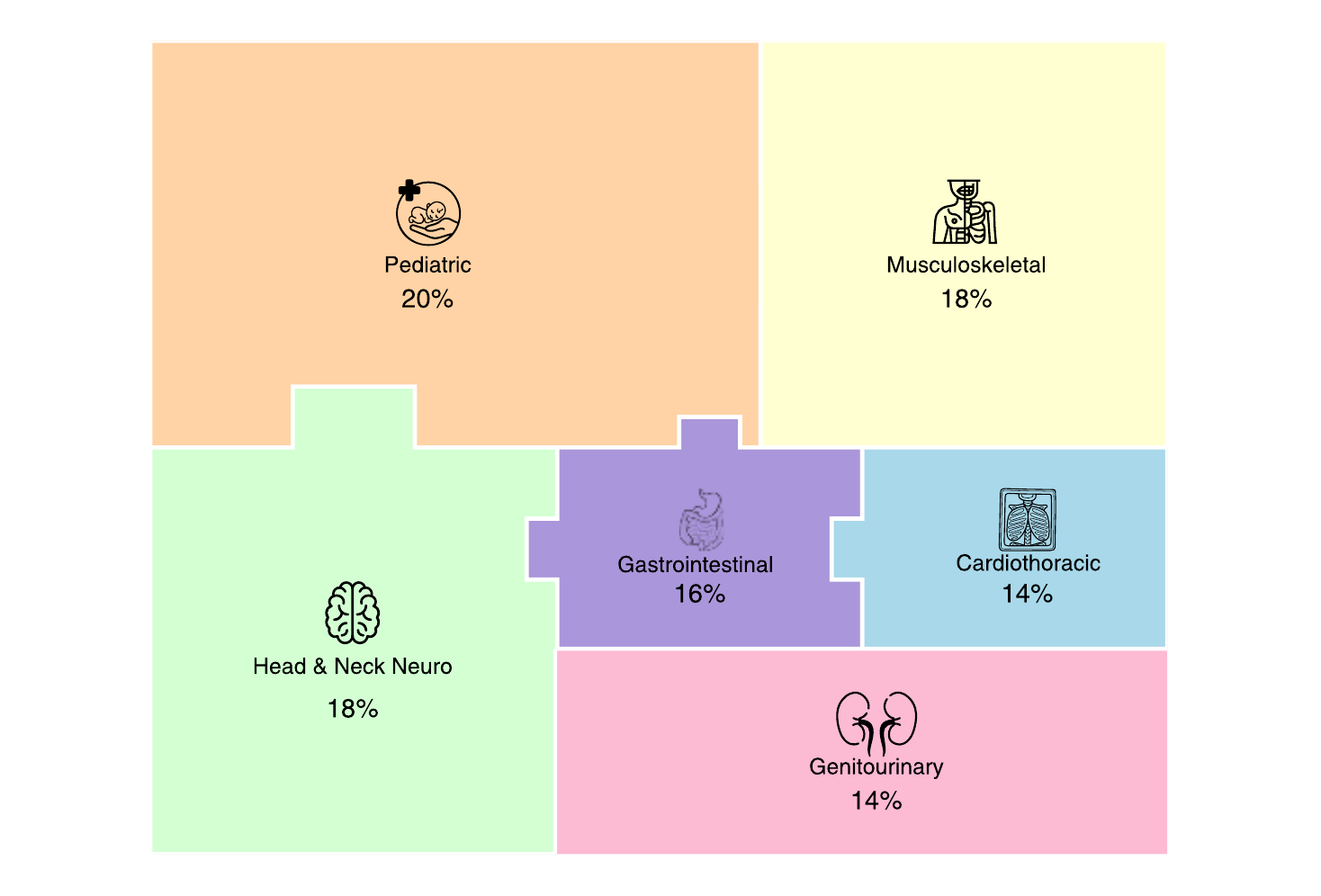}
    \caption{Case distribution by clinical system, illustrating the allocation of spot-diagnosis cases across various anatomical systems within the benchmark dataset.}
\end{figure}

\subsection{Model Evaluation Framework}

\subsubsection{Web Interface Evaluation Protocol}
Five widely used generalist AI models were evaluated through their publicly accessible web interfaces. To ensure reproducibility and minimize bias, three independent evaluations were conducted by different readers for each model. All evaluations followed a standardized protocol to maintain consistency across assessments.

\subsubsection{Data Privacy and Security Measures}
\begin{itemize}
    \item Platform-level data sharing settings were disabled across all interfaces
    \item Temporary chat sessions were utilized where available
    \item All conversation histories were deleted prior to testing subsequent cases
    \item No data retention was permitted during the evaluation process
\end{itemize}

\subsubsection{Evaluated Models}
The following models were assessed in their ``reasoning'' or ``thinking'' modes:
\begin{itemize}
    \item \textbf{OpenAI o3}: Accessed through Web Interface on August 4, 2025
    \item \textbf{Gemini 2.5 Pro}: Accessed through Web Interface on August 5, 2025
    \item \textbf{Grok-4}: Accessed through Web Interface on August 6, 2025
    \item \textbf{Claude Opus 4.1}: Accessed through Web Interface on August 6, 2025
    \item \textbf{OpenAI GPT-5}: Accessed through Web Interface on August 12, 2025
\end{itemize}

\subsubsection{Inter-Reader Reliability Assessment}
To assess consistency in model outputs and evaluation methodology, each model underwent three independent evaluations conducted by different readers. All evaluations used identical case scenarios and standardized prompting protocols under controlled parameter conditions to ensure reproducibility and minimize inter-reader variability.

\subsection{API-Based Evaluation: GPT-5 Reasoning Modes}

\subsubsection{Reasoning Mode Configuration}
GPT-5 was additionally evaluated through its API to assess performance across various reasoning intensities. Three distinct reasoning approaches were tested:
\begin{itemize}
    \item MODE 1: Low effort reasoning (\texttt{effort: "low"})
    \item MODE 2: Medium effort reasoning (\texttt{effort: "medium"})
    \item MODE 3: High effort reasoning (\texttt{effort: "high"})
\end{itemize}

\subsubsection{Technical Parameters}
All API calls utilized the service's default configuration parameters to ensure standardized conditions. The following parameters were maintained at their default values:

\begin{verbatim}
max_tokens = 512, parallel_tool_calling = False, stream = False, response_format = text, 
verbosity = medium, stop_sequence = None, presence_penalty = 0, frequency_penalty = 0, 
previous_response_id = None
\end{verbatim}

Default parameters were maintained throughout all API evaluations as they were not explicitly specified in the evaluation script, ensuring consistency with standard implementation practices.

\subsubsection{Response Latency Measurement}
For GPT-5 API evaluations, response latency was measured using high-precision timing. Before each API call, a timer was initiated using Python's \texttt{time.perf\_counter()} function.  The timers started before each GPT-5 API call and were stopped immediately upon response receipt, and elapsed time was calculated in milliseconds. Latency measurements were recorded for each reasoning effort level (low, medium, high) in dedicated columns within the results dataset to enable computational cost analysis across reasoning modes.

\subsubsection{Prompting Protocol}
All models received an identical standardized prompt that was developed through consensus by two radiologists (one with 5 years of AI research experience, including 2 years in natural language processing and large language models) and optimized specifically for reasoning/thinking modes.

\begin{verbatim}
"You are a board-certified diagnostic radiologist. Given a medical image, 
your task is to provide the single most specific final diagnosis.

Instructions:
- State only the most specific diagnosis, in full words (no abbreviations).
- Do not provide any description, explanation, uncertainty statements, or additional text.
- Output only the diagnosis as a single line - no other information.

Example outputs (for demonstration only):
- Pulmonary tuberculosis
- Von Hippel-Lindau syndrome
- Adrenal hematoma
- Segond fracture

Please review the image and respond as instructed, with only the diagnosis 
in a single line."
\end{verbatim}

\subsection{Scoring and Statistical Analysis}

\subsubsection{Study Population and Case Selection}
All fifty cases were independently interpreted by:
\begin{itemize}
    \item Four board-certified radiologists
    \item Four radiology trainees
    \item Five large language models: OpenAI GPT-5 (hereafter referred to as GPT-5), Gemini 2.5 Pro, OpenAI o3, Grok-4, and Claude Opus 4.1
\end{itemize}

Each reader provided a single most specific diagnosis per case.

\subsubsection{Scoring System}
Diagnoses were graded against reference answers using an ordinal scale.
\begin{itemize}
    \item 1.0: Exact match with reference diagnosis
    \item 0.5: Partially correct differential diagnosis
    \item 0.0: Incorrect diagnosis
\end{itemize}

\subsubsection{Statistical Analysis Framework}
\textbf{Accuracy Analysis.}  
For statistical comparison, scores were aggregated as follows:
\begin{itemize}
    \item Trainee scores (n=4) were averaged within their cohort
    \item Board-certified radiologist scores (n=4) were averaged within their cohort
    \item Model outputs (3 independent runs each) were scored and averaged case-wise for each model
\end{itemize}
This aggregation yielded seven reader cohorts for comparative analysis.

\textbf{Primary Statistical Methods.}  
Overall performance differences were evaluated using the Friedman rank test. When omnibus results were significant, pairwise comparisons were conducted using Wilcoxon signed-rank tests with Holm adjustment.

\textbf{Subgroup Analysis.}  
System-specific and modality-specific subsets contained insufficient observations for reliable statistical inference and were therefore summarized descriptively using mean accuracies with Wilson 95\% confidence intervals.

\subsection{Reliability Analysis}
Reliability assessment preserved all individual scores without aggregation, retaining each model's three raw scores.

\subsubsection{Reliability Metrics}
\begin{itemize}
    \item Quadratic-weighted kappa coefficients were calculated for every pair of runs
    \item Two-way random-effects intraclass correlation coefficients [ICC(2,1)] were computed across triplicate model scores for each AI system
\end{itemize}

\subsection{Software Implementation}
All statistical procedures were performed in R version 4.5.0 using:
\begin{itemize}
    \item Base functions for non-parametric tests
    \item \texttt{lme4} package for mixed-effects modeling
    \item \texttt{irr} package for reliability statistics
\end{itemize}

\section{Results}

\subsection{Overall Performance Comparison}
Board-certified radiologists achieved the highest mean diagnostic accuracy at 83\% (95\% CI: 75--90\%) and significantly outperformed every comparison group (Friedman $\chi^{2}$ = 336, Kendall W = 0.56, $p < 1 \times 10^{-64}$). Radiology trainees followed with 45\% accuracy (95\% CI: 39--52\%) and remained superior to all AI models tested.

Among all AI models (with Thinking or Reasoning modes toggled on), GPT-5 demonstrated the highest performance at 30\% accuracy (95\% CI: 20--42\%), followed closely by Gemini 2.5 Pro at 29\% (95\% CI: 19--39\%). OpenAI o3 achieved 23\% accuracy (95\% CI: 14--33\%), while Grok-4 reached 12\% (95\% CI: 6--19\%), and Claude Opus 4.1 performed poorly at 1\% accuracy (95\% CI: 0--3\%). Each AI model's accuracy was significantly lower than the trainee cohort (Holm-adjusted Wilcoxon $p \leq 0.010$) and substantially below the radiologist benchmark, confirming a persistent human--AI performance gap in spot-diagnosis tasks. Further details are given in Table 3.

\begin{table}[ht]
\section*{Table 3. Overall diagnostic accuracy across reader groups}
\centering
\begin{tabular}{p{0.35\linewidth} p{0.25\linewidth} p{0.25\linewidth}}
\toprule
\textbf{Group} & \textbf{Mean Accuracy} & \textbf{95\% CI} \\
\midrule
\textbf{Board-certified radiologists (n=4)} & \textbf{0.83} & \textbf{0.75--0.90} \\
Radiology trainees (n=4) & 0.45 & 0.39--0.52 \\
GPT-5 & 0.30 & 0.20--0.42 \\
Gemini 2.5 Pro & 0.29 & 0.19--0.39 \\
OpenAI o3 & 0.23 & 0.14--0.33 \\
Grok-4 & 0.12 & 0.06--0.19 \\
Claude Opus 4.1 & 0.01 & 0.00--0.03 \\
\botrule
\end{tabular}

\vspace{0.5em}
\noindent\footnotesize\emph{Mean accuracy represents the proportion of exact and partial correct responses across 50 radiology spot-diagnosis cases. Confidence intervals calculated using Wilson method. Friedman $\chi^2$ = 336.0, df = 6, $p < 1\times10^{-64}$; Kendall's W = 0.56 (large effect size). Scoring: exact match (1.0), partial correct (0.5), incorrect (0.0).}

\end{table}

\begin{figure}[htbp]
    \centering
    \includegraphics[width=0.9\linewidth]{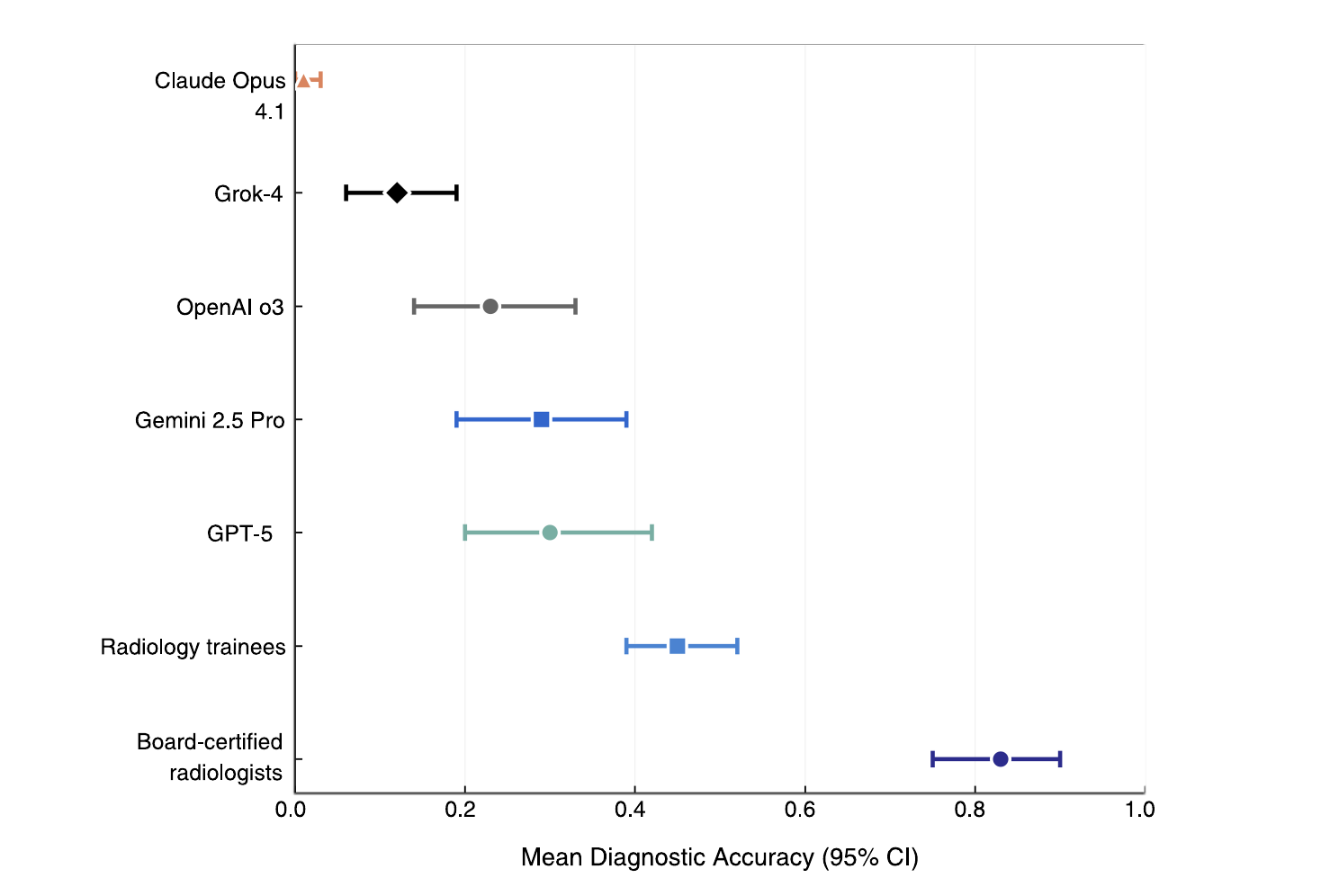}
    \caption{Overall diagnostic accuracy across reader groups. This figure presents the mean diagnostic accuracy with 95\% confidence intervals for board-certified radiologists, radiology trainees, and five frontier AI models (GPT-5, Gemini 2.5 Pro, OpenAI o3, Grok-4, and Claude Opus 4.1) on 50 challenging radiology spot-diagnosis cases.}
\end{figure}

\subsection{Performance by Imaging Modality}
Performance varied across imaging modalities, with board-certified radiologists maintaining superiority across all modalities.

\begin{itemize}
    \item \textbf{Computed Tomography (24 cases):} Radiologists led with 79\% accuracy, trainees achieved 57\%. Among AI systems, Gemini 2.5 Pro performed best at 29\%, followed by GPT-5 at 22\%, OpenAI o3 at 19\%, Grok-4 at 8\%, and Claude Opus 4.1 at 1\%.
    \item \textbf{Magnetic Resonance Imaging (13 cases):} Radiologists demonstrated 98\% accuracy, trainees 58\%. GPT-5 topped AI models at 45\%, followed by Gemini 2.5 Pro (35\%), OpenAI o3 (33\%), Grok-4 (23\%), and Claude Opus 4.1 (0\%).
    \item \textbf{Plain Radiography (13 cases):} Radiologists scored 89\% accuracy, trainees 53\%. GPT-5 led AI systems at 31\%, Gemini 2.5 Pro and OpenAI o3 each at 22\%, Grok-4 at 8\%, and Claude Opus 4.1 at 3\%.
\end{itemize}

AI models performed best on MRI cases, with GPT-5 achieving 45\% accuracy compared to 22\% on CT and 31\% on radiographs. Further details are provided in Table 4.

\begin{table}[ht]
\section*{Table 4. Diagnostic accuracy by imaging modality}
\centering
\setlength{\tabcolsep}{1pt}
\renewcommand{\arraystretch}{2}

\begin{tabularx}{\linewidth}{
    >{\raggedright\arraybackslash}p{0.12\linewidth} 
    *{7}{>{\centering\arraybackslash}p{0.12\linewidth}} 
}
\toprule
\textbf{Modality} &
\textbf{Radiologists} &
\textbf{Trainees} &
\textbf{GPT-5} &
\makecell{\textbf{Gemini 2.5}\\\textbf{Pro}} &
\makecell{\textbf{OpenAI}\\\textbf{o3}} &
\textbf{Grok-4} &
\makecell{\textbf{Claude Opus}\\\textbf{4.1}} \\
\midrule
CT (n=24) &
\textbf{0.79} (0.60--0.91) &
0.57 (0.39--0.76) &
0.22 (0.12--0.35) &
0.29 (0.18--0.41) &
0.19 (0.09--0.32) &
0.08 (0.02--0.17) &
0.01 (0.00--0.05) \\
MRI (n=13) &
\textbf{0.98} (0.77--1.00) &
0.58 (0.36--0.82) &
0.45 (0.25--0.67) &
0.35 (0.17--0.56) &
0.33 (0.15--0.54) &
0.23 (0.08--0.44) &
0.00 (0.00--0.08) \\
Radiograph (n=13) &
\textbf{0.89} (0.67--0.99) &
0.53 (0.29--0.77) &
0.31 (0.14--0.52) &
0.22 (0.08--0.41) &
0.22 (0.08--0.41) &
0.08 (0.01--0.22) &
0.03 (0.00--0.12) \\
\botrule
\end{tabularx}

\vspace{0.5em}
\noindent\footnotesize\emph{Mean proportion of correct classifications with Wilson 95\% confidence intervals. CT = computed tomography; MRI = magnetic resonance imaging.}
\end{table}

\begin{figure}[htbp]
    \centering
    \includegraphics[width=0.9\linewidth]{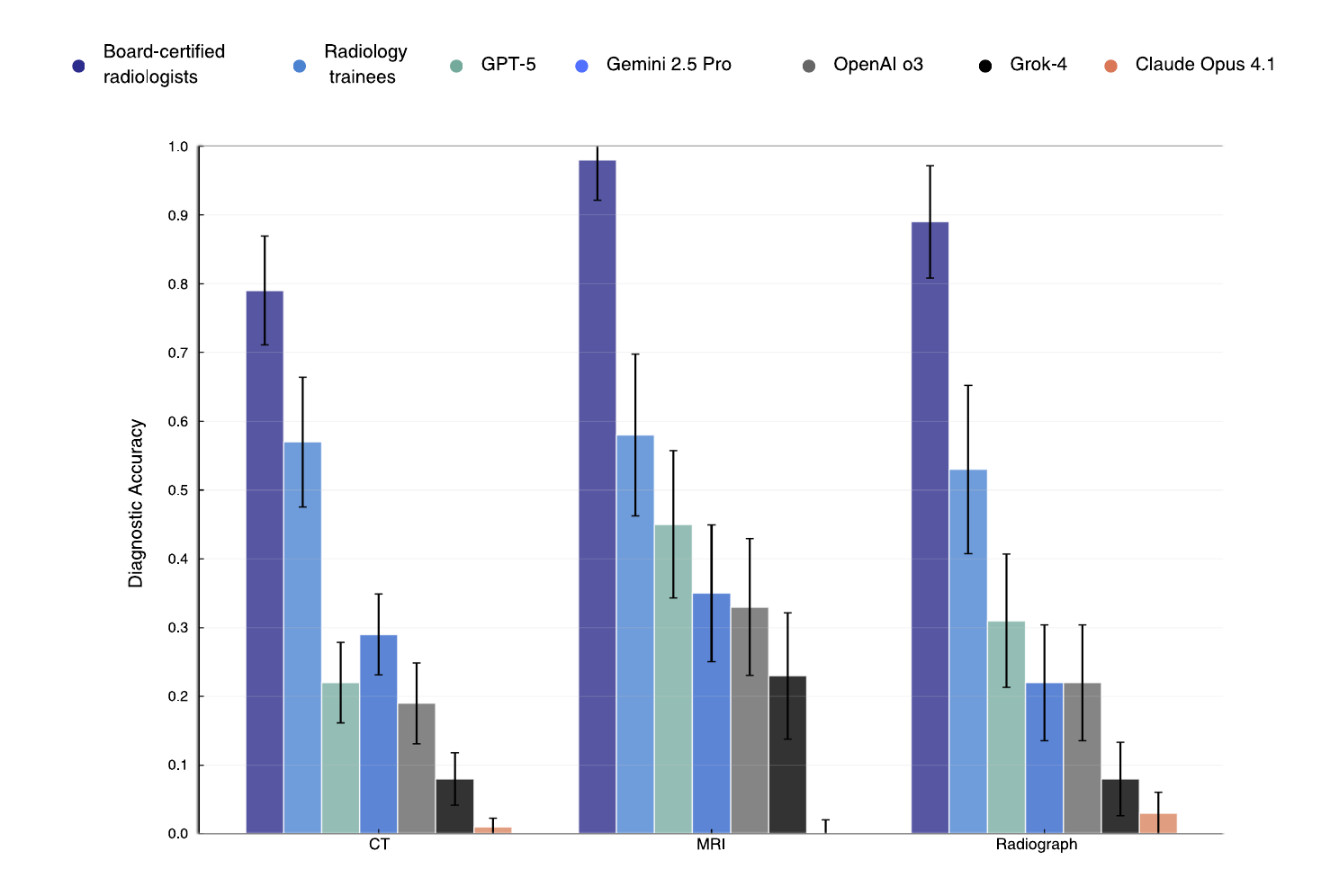}
    \caption{Detailed modality-specific diagnostic accuracy. This figure expands on the modality-level comparison, showing mean diagnostic accuracies with 95\% confidence intervals for radiologists, trainees, and five frontier AI models (GPT-5, Gemini 2.5 Pro, OpenAI o3, Grok-4, and Claude Opus 4.1) across CT, MRI, and Radiography.}
\end{figure}

\begin{figure}[htbp]
    \centering
    \includegraphics[width=0.9\linewidth]{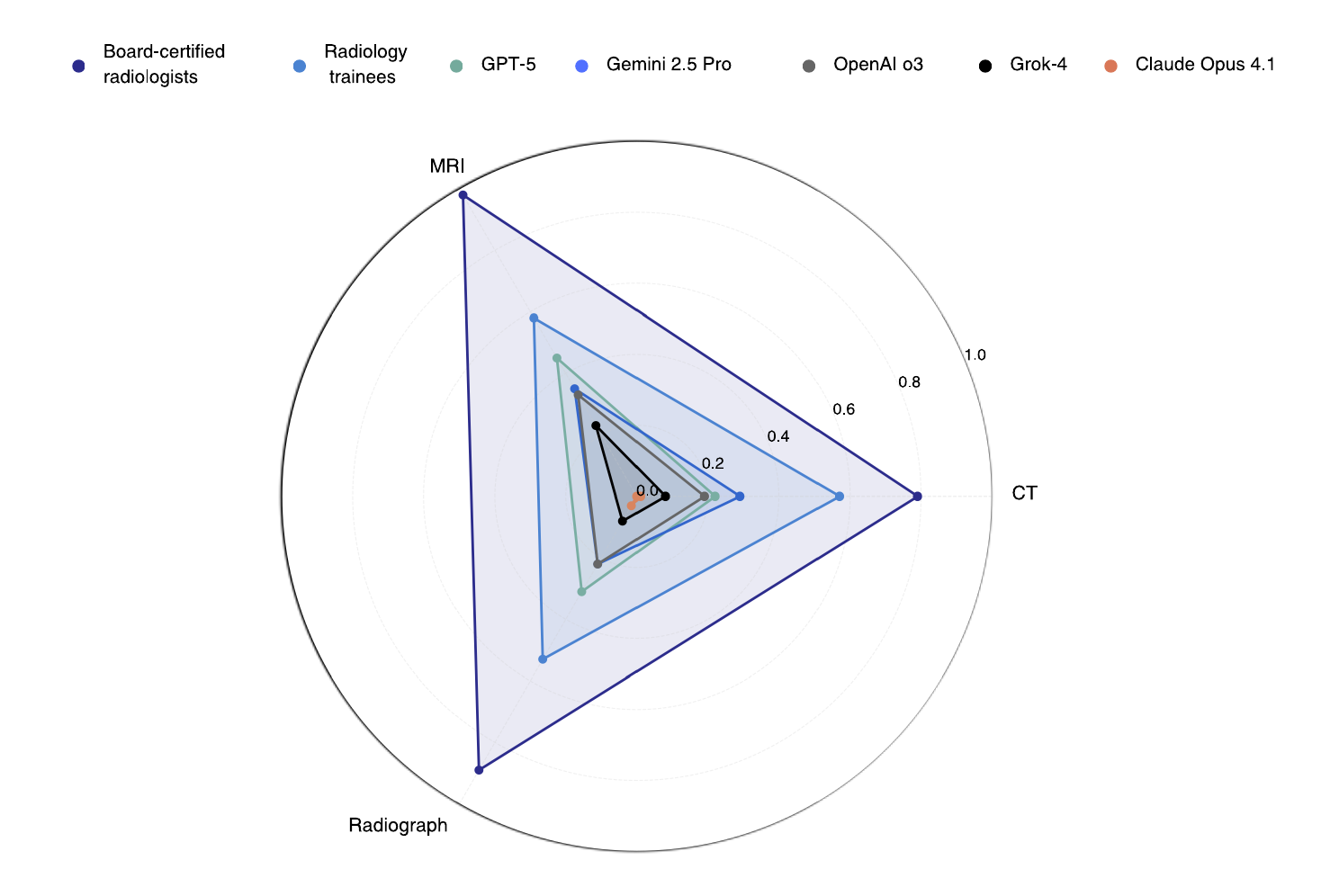}
    \caption{Modality-specific diagnostic accuracy for radiologists, trainees, and large-language models. This radar plot illustrates the diagnostic accuracy of board-certified radiologists, radiology trainees, and five frontier AI models (GPT-5, Gemini 2.5 Pro, OpenAI o3, Grok-4, and Claude Opus 4.1) across three imaging modalities (CT, MRI, and Radiograph).}
\end{figure}

\subsection{Performance by System}
Board-certified radiologists maintained superior performance across all anatomical systems, with at least a 0.25 absolute accuracy margin over trainees and $\geq$0.50 margin over the best-performing AI models. Trainees consistently ranked second across all systems. GPT-5 and Gemini 2.5 Pro alternated as leading AI models but never surpassed trainee performance. The overall performance hierarchy was preserved within each subspecialty. Details are given in Table 5.

\begin{table}[ht]
\section*{Table 5. System-specific diagnostic accuracy for radiologists, trainees and large-language models}
\centering
\setlength{\tabcolsep}{1pt}
\renewcommand{\arraystretch}{2}

\begin{tabularx}{\linewidth}{
  >{\raggedright\arraybackslash}p{0.15\linewidth} 
  *{7}{>{\centering\arraybackslash}p{0.12\linewidth}} 
}
\toprule
\textbf{System (n)} &
\textbf{Radiologists} & \textbf{Trainees} &
\textbf{GPT-5} &
\makecell{\textbf{Gemini 2.5}\\\textbf{Pro}} &
\makecell{\textbf{OpenAI}\\\textbf{o3}} &
\textbf{Grok-4} &
\makecell{\textbf{Claude Opus}\\\textbf{4.1}} \\
\midrule
Cardiothoracic (n=9) &
\textbf{0.80} (0.63--0.91) & 0.54 (0.37--0.70) & 0.32 (0.18--0.48) & 0.31 (0.17--0.47) & 0.24 (0.12--0.40) & 0.10 (0.03--0.24) & 0.02 (0.00--0.09) \\
Gastrointestinal (n=9) &
\textbf{0.84} (0.66--0.95) & 0.56 (0.39--0.72) & 0.30 (0.17--0.46) & 0.32 (0.19--0.48) & 0.23 (0.12--0.38) & 0.11 (0.04--0.24) & 0.01 (0.00--0.09) \\
Genitourinary (n=8) &
\textbf{0.82} (0.64--0.94) & 0.55 (0.38--0.71) & 0.28 (0.15--0.44) & 0.30 (0.17--0.46) & 0.22 (0.11--0.37) & 0.09 (0.02--0.22) & 0.02 (0.00--0.10) \\
Musculoskeletal (n=10) &
\textbf{0.81} (0.63--0.92) & 0.53 (0.36--0.69) & 0.29 (0.16--0.45) & 0.31 (0.18--0.47) & 0.23 (0.12--0.39) & 0.12 (0.04--0.25) & 0.02 (0.00--0.10) \\
Neuro (n=8) &
\textbf{0.90} (0.74--0.98) & 0.59 (0.42--0.74) & 0.33 (0.19--0.49) & 0.30 (0.17--0.46) & 0.25 (0.13--0.41) & 0.13 (0.05--0.26) & 0.01 (0.00--0.09) \\
Paediatric (n=6) &
\textbf{0.88} (0.70--0.97) & 0.52 (0.35--0.68) & 0.27 (0.14--0.43) & 0.28 (0.15--0.44) & 0.21 (0.10--0.36) & 0.09 (0.02--0.22) & 0.00 (0.00--0.08) \\
\botrule
\end{tabularx}

\vspace{0.5em}
\noindent\footnotesize\emph{Mean proportion of correct classifications with Wilson 95\% confidence intervals across anatomical systems. Some confidence intervals are wide due to limited case numbers per system.}
\end{table}

\begin{figure}[htbp]
    \centering
    \includegraphics[width=0.9\linewidth]{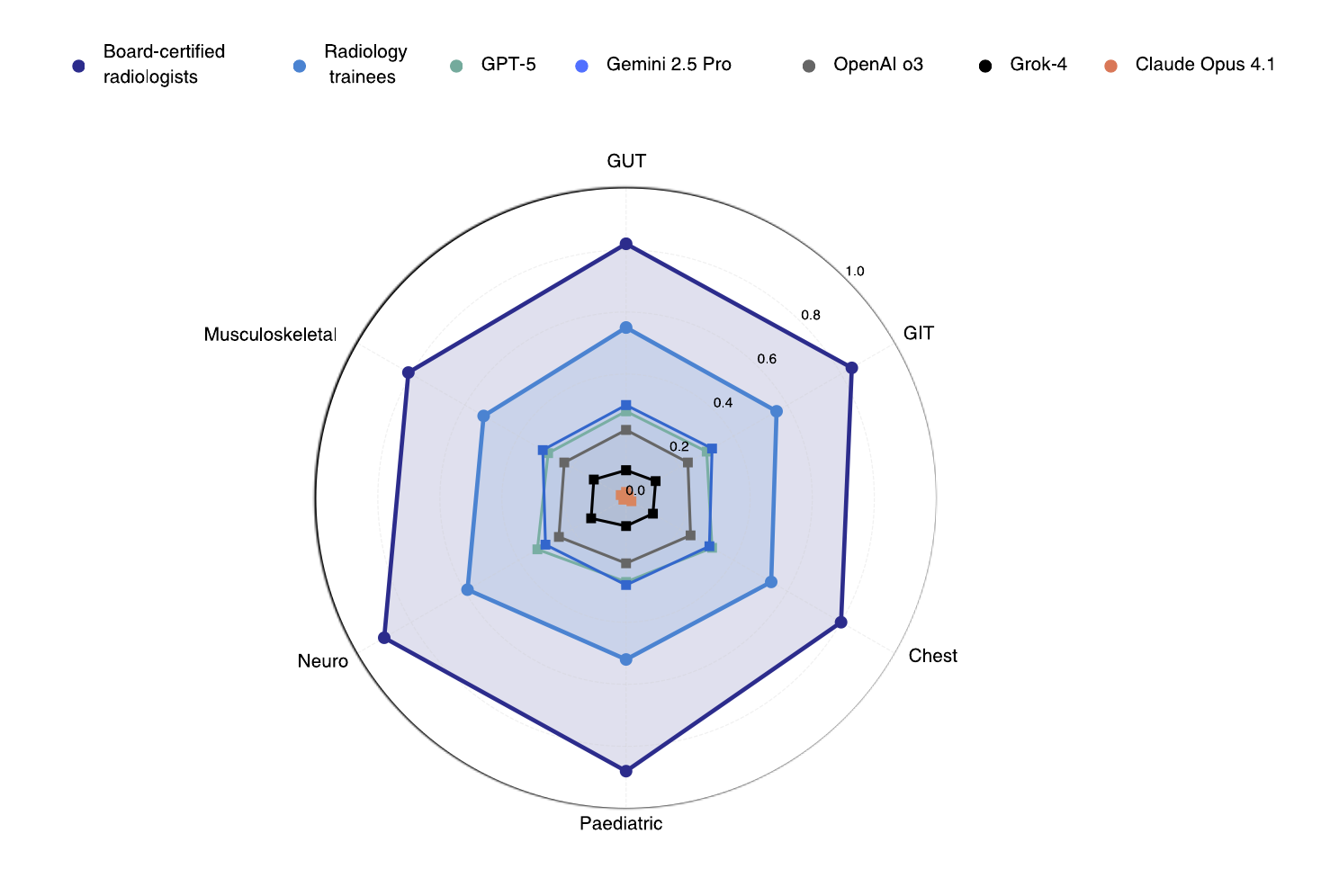}
    \caption{System-specific diagnostic accuracy for radiologists, trainees, and large-language models. This radar plot illustrates the diagnostic accuracy of board-certified radiologists, radiology trainees, and five frontier AI models (GPT-5, Gemini 2.5 Pro, OpenAI o3, Grok-4, and Claude Opus 4.1) across anatomical systems (Musculoskeletal, Gastrointestinal/GIT, Genitourinary/GUT, Cardiothoracic/Chest, Neuro, Paediatrics).}
\end{figure}

\begin{figure}[htbp]
    \centering
    \includegraphics[width=0.9\linewidth]{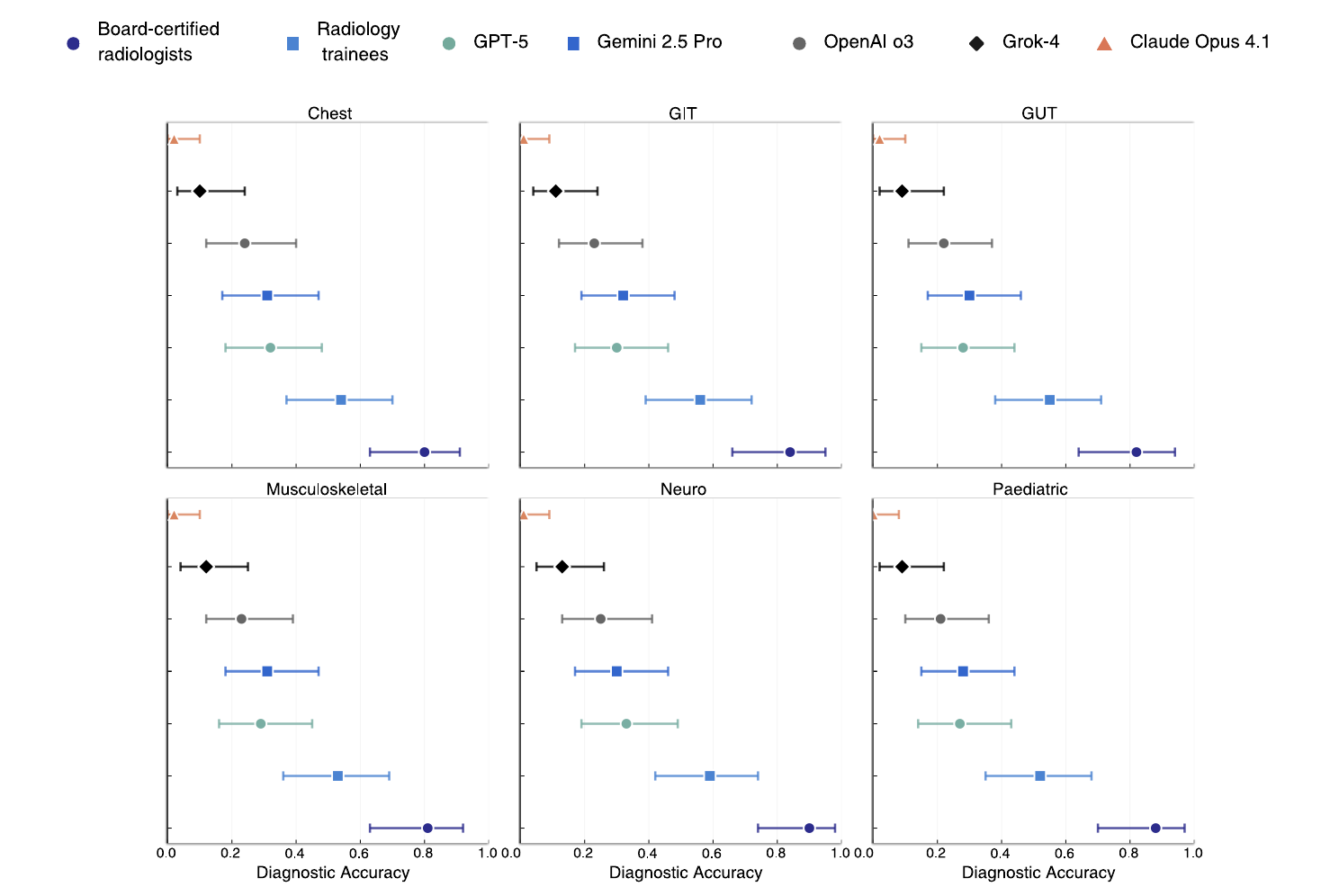}
    \caption{Comparative diagnostic accuracy of human experts (board-certified radiologists and trainees) and frontier AI models (GPT-5, Gemini 2.5 Pro, OpenAI o3, Grok-4, and Claude Opus 4.1) across various anatomical systems.}
\end{figure}

\subsection{GPT-5 Reasoning Mode Analysis}
GPT-5's reasoning mode adjustments through API yielded minimal performance differences across effort levels: Low effort = 25\%, Medium effort = 25\%, High effort = 26\%. The largest observed improvement was 1 percentage point from Low to High mode, with all three settings remaining substantially below both trainee and expert benchmarks (Holm-adjusted $p = 1.00$ for all pairwise comparisons with human readers). This minimal accuracy gain came at substantial computational cost.

\subsection{Response Latency Analysis}
Response latency increased significantly with reasoning effort level:
\begin{itemize}
    \item Low effort: Mean latency 10,475 ms (SD: 5,030 ms, n = 50)
    \item Medium effort: Mean latency 28,849 ms (SD: 14,561 ms, n = 50)
    \item High effort: Mean latency 65,584 ms (SD: 33,469 ms, n = 50)
\end{itemize}

High-effort tasks required over 6$\times$ longer than low-effort mode (65.6 vs 10.5 seconds). Variability was also highest in high-effort mode, indicating unpredictable delays that could significantly impact clinical workflow.

\subsection{Consistency Analysis}
GPT-5 showed the strongest repeatability with quadratic-weighted kappa values of 0.59–0.73 (mean $\kappa$ = 0.64) and ICC(2,1) = 0.64 (95\% CI: 0.50–0.76), indicating substantial agreement. OpenAI o3 showed comparable consistency ($\kappa$ = 0.54–0.75; ICC = 0.61, 95\% CI: 0.46–0.74). Gemini 2.5 Pro and Grok-4 achieved moderate agreement levels ($\kappa$ = 0.47–0.61 and 0.27–0.62; ICC = 0.54 and 0.41, respectively). Claude Opus 4.1 demonstrated poor reproducibility ($\kappa \approx -0.02$–0.00; ICC $\approx 0$). Full details are provided in Table 6.

\begin{table}[ht]
\section*{Table 6. Consistency and repeatability of AI systems across 3 runs}
\centering
\setlength{\tabcolsep}{1pt}

\begin{tabularx}{\linewidth}{
  >{\raggedright\arraybackslash}p{0.20\linewidth}
  >{\centering\arraybackslash}p{0.16\linewidth}
  >{\centering\arraybackslash}p{0.12\linewidth}
  >{\raggedright\arraybackslash}p{0.18\linewidth}
  >{\centering\arraybackslash}p{0.12\linewidth}
  >{\centering\arraybackslash}p{0.12\linewidth}
}
\toprule
\textbf{Model} & \textbf{$\kappa$ range} & \textbf{Mean $\kappa$} & \textbf{Interpretation} & \textbf{ICC(2,1)} & \textbf{95\% CI} \\
\midrule
\textbf{GPT-5} &\textbf{ 0.59--0.73} & \textbf{0.64} & \textbf{Substantial} & \textbf{0.64} & \textbf{0.50--0.76} \\
OpenAI o3        & 0.54--0.75 & 0.61 & Substantial & 0.61 & 0.46--0.74 \\
Gemini 2.5 Pro   & 0.47--0.61 & 0.53 & Moderate    & 0.54 & 0.37--0.68 \\
Grok-4           & 0.27--0.62 & 0.41 & Moderate    & 0.41 & 0.24--0.59 \\
Claude Opus 4.1  & --0.02--0.00 & --0.01 & Poor       & $\approx 0$ & --0.15--0.17 \\
\bottomrule
\end{tabularx}

\vspace{0.5em}
\noindent\footnotesize\emph{$\kappa$ = quadratic-weighted kappa; ICC = intraclass correlation coefficient. $\kappa$ confidence intervals omitted because each estimate derived from a single 50-case contingency table; ICCs provide formal interval estimates.}
\end{table}

\section{Visual Reasoning Error Analysis}

To understand the diagnostic failure modes underlying the quantitative performance gaps, we conducted a qualitative analysis of reasoning traces generated by the AI models. The systematic evaluation of diagnostic accuracy in vision language models requires a structured framework for characterizing failure modes. Drawing from established radiology error taxonomies, particularly the work of Kim and Mansfield \cite{kim2014fool} and the comprehensive bias analysis by Onder et al. \cite{onder2021errors}, we propose a taxonomy specifically adapted for the analysis of reasoning traces generated by vision language models operating on single medical images.

\subsection{Framework Development}
Our error classification system was developed through iterative review sessions involving board-certified radiologists and cognitive psychologists. The framework necessarily focuses on errors observable within generated reasoning text from single-image analysis without any clinical history, prior examinations or real-time clinical consultation. This constrained setting excludes several error categories established in radiology literature including technique-related errors, procedure-related complications, history-dependent misinterpretations and inter-physician communication failures but enables focused analysis of pure visual reasoning capabilities of AI models. We also illustrate examples of each error type through reasoning traces from GPT-5, the best performing model.

\subsection{Proposed Error Classification System}
The taxonomy organizes diagnostic reasoning errors into three primary categories: perceptual, interpretive, and communication errors. Additionally, cognitive bias patterns observed in reasoning traces are included as modifiers that may influence error manifestation (Table~7).

\begin{table}[ht]
\section*{Table 7. Taxonomy of visual reasoning errors for single-image diagnostic tasks}
\centering
\setlength{\tabcolsep}{1pt}
\renewcommand{\arraystretch}{1.8}

\begin{tabularx}{\linewidth}{>{\raggedright\arraybackslash}p{0.28\linewidth} >{\raggedright\arraybackslash}X}
\toprule
\textbf{Category} & \textbf{Subtypes} \\
\midrule
\textbf{Perceptual errors} & Under-detection \newline Over-detection \newline Mislocalization \\
\textbf{Interpretive errors} & Misinterpretation or misattribution of findings \newline Incomplete reasoning or premature diagnostic closure \\
\textbf{Communication errors} & Findings–summary discordance \\
\textbf{Cognitive bias modifiers} & Confirmation/Anchoring bias \newline Availability bias \newline Inattentional bias \newline Framing effects \\
\botrule
\end{tabularx}
\end{table}

\subsubsection{Perceptual Errors}

\textbf{Under-detection} occurs when reasoning traces fail to identify or describe visible pathological findings present in the image. This parallels one of the most prevalent error types in clinical radiology \cite{brady2017error}. In a representative case demonstrating left ureterocele, GPT-5 failed to identify dilatation/ballooning of distal ureter prolapsing into the urinary bladder on IVU despite its clear visibility to radiologists.

\textbf{Over-detection} captures confident identification of pathological findings not visually supported by the image evidence, potentially reflecting model hallucination tendencies. In the same ureterocele case, the model not only failed to identify the primary pathology but also confidently reported multiple cysts with spider-like pattern in bilateral kidneys despite the normal appearance of the kidneys, which led to the incorrect diagnosis of Autosomal Dominant Polycystic Kidney Disease.

\textbf{Mislocalization} represents correct identification of pathological patterns but incorrect spatial attribution to wrong anatomical locations, sides, or compartments. In a case of a right atrium hydatid cyst, GPT-5 accurately identified a cystic lesion within the mediastinum, but failed to precisely localize it to the right atrium.

\subsubsection{Interpretive Errors}
\textbf{Misinterpretation or misattribution of findings} occurs when visual patterns are correctly identified but incorrectly linked to pathophysiological processes or differential diagnoses. In a case of acromioclavicular dislocation, GPT-5 attempted to ascertain the relationship between the distal clavicle and acromion, and successfully identified clavicular elevation. However, the final conclusive diagnosis provided by the model was posterior shoulder dislocation.

\textbf{Incomplete reasoning or premature diagnostic closure} captures instances where initial diagnostic impressions are accepted without adequate consideration of alternatives, mirroring the ``premature closure'' bias documented in human diagnostic reasoning \cite{croskerry2003cognitive}. In one case of Joubert's syndrome, GPT-5 concluded Central Pontine Myelinolysis based solely on pontine involvement and prematurely ended the diagnostic reasoning stating “no further explanations needed”.

\subsubsection{Communication Errors}
\textbf{Findings-summary discordance} identifies internal inconsistencies within reasoning traces, where detailed observations contradict final diagnostic impressions. This error type raises questions about reasoning chain stability in autoregressive language models. GPT-5 was presented with an abnormal chest X-ray, the model identified findings such as slender heart with a low cardiothoracic ratio, depressed diaphragms demonstrating hyperinflation. However, the model finally concluded it as a normal chest radiograph.

\subsection{Cognitive Bias Modifiers}
Four cognitive bias patterns are observed to influence diagnostic reasoning:

\textbf{Confirmation bias / anchoring bias} was observed as early fixation on initial diagnostic hypotheses with subsequent favouring of supporting evidence, despite identifying contradictory findings. In a case of proximal femoral deficiency, the model initially assumed bilateral femoral displacements. Although its intermediate reasoning briefly identified features of proximal femoral deficiency in one of the limbs, it ultimately returned to its initial diagnosis and finally concluded it as developmental dysplasia of the hip.

\textbf{Availability bias} manifested as apparent overweighting of diagnoses potentially over-represented in training data.  GPT-5 was presented with multiple abnormal chest X-rays/CT with diagnosis varying from pulmonary arteriovenous malformation, Scimitar syndrome etc., the model exhibited a propensity for diagnosis of these cases predominantly as acute conditions such as pneumonia, pulmonary embolism, probably attributable to abundance of such cases in available public datasets on which these models were trained.

\textbf{Inattentional bias} presented as neglect of relevant anatomical regions or findings despite comprehensive visual analysis capabilities. In a case of Portal Hypertension with esophageal varices, shrunken liver and splenomegaly, GPT-5 only focussed on artefactual hypodensity in the spleen; however, it neglected the hepatic and the lower esophageal anatomy failing to recognize cirrhotic findings in the liver and the esophageal varices.

\textbf{Framing effects} occurred when prompt structure appeared to bias interpretation toward specific diagnostic categories. In a case of tarsal coalition which included both talocalcaneal and calcaneonavicular coalitions, the model identified both talocalcaneal and calcaneonavicular coalitions, but ultimately provided a final diagnosis of calcaneonavicular coalition, likely because the prompt requested a single, specific diagnosis.

These patterns were identified through qualitative analysis of reasoning traces and represent observed tendencies rather than definitively established mechanisms. This is a reminder that even in our attempts to understand artificial intelligence, we remain fundamentally in the dark about the computational mysteries unfolding within these black boxes. We can observe what these models do wrong, but the deeper question of why they fail in these ways remains as opaque as consciousness itself.

\section{Discussion}

\subsection{Performance Gap Analysis}
Our findings demonstrate that frontier generalist multimodal AI systems remain substantially below the diagnostic accuracy of both radiology trainees and board-certified experts on complex radiological cases. The 30\% accuracy achieved by the best-performing model GPT-5 on difficult cases, compared to 83\% for board-certified radiologists, highlights both the progress made and the substantial development needed for clinical deployment. 

Recent evaluations provide important context for these performance gaps. Wang et al.\ (2025) found that GPT-5 achieved state-of-the-art accuracy across medical question-answering benchmarks, even surpassing ``pre-licensed medical experts'' by 24\% in reasoning ability on structured evaluations \cite{wang2025gpt5}. Similarly, Hou et al.\ (2025) reported that OpenAI's o1 model scored 59\% on RSNA Case of the Day challenges, statistically matching expert radiologists on exam-format questions \cite{hou2025oneyear}. These structured evaluation successes contrast sharply with our 30\% accuracy finding, highlighting the substantial gap between performance on formatted exam questions versus complex, real-world spot diagnoses.

\subsection{Evaluation Methodology Considerations}
The observed variability across runs highlights the challenge of evaluating models through public-facing web interfaces, where hidden parameters and sudden version updates may influence reproducibility. Consistency was evaluated for all models through multiple repeat runs, though fixed decoding parameters could not be guaranteed across consumer platforms.

While platform settings were configured to minimize data retention, the lack of full transparency in consumer systems raises questions about auditability and long-term governance of clinical AI usage. The partial credit scheme, though designed to capture near-miss diagnoses or broad applicability, highlights the need for standardized ontology-based frameworks for evaluating AI reasoning in clinical diagnosis.

These methodological concerns align with independent evaluations by Brin et al.\ (2025), who found that GPT-4V (GPT-4 Vision) achieved only 35\% pathology recognition accuracy in emergency radiology cases and exhibited high hallucination rates, consistent with the perceptual errors and over-detection patterns identified in our taxonomy \cite{brin2025assessing}.

\subsection{Spectrum Bias Considerations}
This study intentionally focused on challenging, high-complexity radiological cases to probe the upper limits of diagnostic capability. Our pilot dataset of 50 cases was deliberately constructed to test model limits rather than to represent routine case mixes. While this approach introduces spectrum bias, it serves to evaluate AI system robustness under challenging conditions where expert judgment is most critical and where diagnostic errors carry the highest clinical consequences.

\subsection{Technical Insights}
The minimal improvement observed across GPT-5's reasoning modes indicates that extended deliberation provides limited diagnostic benefit, despite substantial computational costs. The six-fold increase in response time for high-effort processing without corresponding accuracy gains suggests that current ``reasoning'' capabilities may not translate effectively to improved diagnostic performance in medical imaging.

\subsection{Reliability and Consistency Implications}
The consistency analysis revealed variability in few of the model outputs, with even the best-performing system achieving only substantial agreement. The observed inconsistency in responses further demonstrates reliability concerns for clinical deployment. In clinical contexts, reproducibility of outputs for identical inputs is essential for maintaining clinician trust and ensuring consistent patient care.

\subsection{Error Pattern Insights}
The systematic error patterns identified through our qualitative analysis provide insights for improving medical AI systems. The prevalence of perceptual errors, particularly under-detection, suggests fundamental limitations in current vision language architectures for medical image analysis that may not be addressed solely through increased model scale or reasoning capabilities. 

Our systematic error categorization is consistent with independent observations by Brin et al.\ (2025), who similarly documented high rates of hallucinated findings and spatial mislocalization errors in GPT-4V's radiological interpretations, suggesting these failure modes are prevalent across non-reasoning and reasoning frontier multimodal models \cite{brin2025assessing}.

\subsection{Clinical Safety Implications}
These results highlight the need for cautious deployment of generalist AI systems in medical contexts either by clinicians or by patients. The substantial performance gaps, combined with evidence of increasing clinical adoption without institutional oversight, raise important safety concerns about over-reliance on such tools in high-stakes diagnostic scenarios.

\subsection{Limitations}

\subsubsection{Methodological Limitations}
This work represents a pilot study ($n=50$ cases). As such, the sample size limits the statistical power of subgroup analyses, and results, especially system-wise or modality-wise, should be interpreted as exploratory. The small human comparator groups ($n=4$ each in the radiologist and the trainee group) constrain characterization of performance variability. Larger, more diverse datasets will be essential to validate and generalize these findings.

\subsubsection{Evaluation Platform Constraints}
Evaluations through public web interfaces are subject to potential version drift and backend changes. Hidden sampling parameters (e.g., temperature, top-p) may vary and outputs may be affected by provider-side A/B testing. In our study, exact model identifiers and timestamps were logged, but sampling controls could not be standardized.

Platform-level data retention controls were disabled where available, yet these measures remain best-effort. Residual retention of metadata or interaction logs cannot be fully excluded.

\subsubsection{Scoring and Assessment Limitations}
Scoring of partial matches involved subjective judgment. Although predefined rules and independent adjudication were applied with consensus decision by radiologists wherever in doubt, some residual subjectivity may be inherent to ontology mapping of radiological diagnoses.

\subsubsection{Dataset and Generalizability Constraints}
The deliberate spectrum bias toward challenging cases limits generalizability to routine clinical practice, while the focus on single-image spot diagnoses without clinical history may not reflect integrated clinical decision-making. The proposed error taxonomy was developed by radiologists and cognitive psychologists from limited sample analysis and may need subclassification of failure modes in larger datasets, a work in progress.

Dataset reproducibility is limited by our current inability to make the evaluation set public to prevent model training contamination. Finally, evaluation was performed on specific model versions available at specific dates. Given the rapid pace of model iteration, these results may not reflect future system performance.

\subsection{Future Directions}
Priority development areas for frontier labs may include improved detection of subtle findings in complex and difficult cases, more meaningful integration of imaging and clinical reasoning, improved response consistency and determinism and specialized fine-tuned models for radiological and medical applications. Alternatively, we believe that these challenges may be best addressed by partnering with dedicated radiology AI startups, a natural division of labor that maintains competitive diversity in the field while ensuring that domain-specific expertise and innovation opportunities remain distributed across multiple smaller entities rather than concentrated within a limited number of large-scale generalist platforms.

\subsection{Regulatory and Implementation Considerations}
Our findings support the need for regulatory frameworks that require model evaluation on high-complexity cases, rather than relying solely on performance metrics from standard datasets. Safe clinical implementation should mandate expert human oversight and clear disclosure of model limitations on challenging benchmarks like ours, particularly in patient-facing contexts.

\section{Conclusion}
This evaluation of frontier generalist multimodal LLMs, including VLMs with ``reasoning'' modes, on our novel \textit{Radiology’s Last Exam (RadLE) v1} dataset comprising challenging radiological cases reveals substantial performance gaps relative to expert radiologists and trainees. While these models display impressive general capabilities, they remain unsuitable for autonomous or reliable use in complex diagnostic settings. The best-performing model, GPT-5, achieved only 30\% accuracy compared with 83\% for board-certified radiologists and 45\% for trainees, with only substantial but not high repeatability and negligible gains from higher-effort reasoning despite considerable latency costs, highlighting both progress and the development still required before dependable clinical deployment.

Beyond quantitative metrics, our systematic analysis of reasoning traces yielded a concise taxonomy of visual reasoning errors by VLMs. Our framework encompassing perceptual errors (under-detection, over-detection, mislocalization), interpretive errors (misinterpretation, premature closure) and communication errors (findings–summary discordance), alongside cognitive bias patterns, provides a structured basis for understanding current model failures and guiding mitigation strategies. By pairing a challenging benchmark with an explicit error taxonomy, we offer a practical template for tracking progress, shaping evidence-based policies for AI use in radiology, and aligning future model development with the realities of expert-level diagnostic practice.

\backmatter

\section*{Declarations}

\textbf{Funding:} None. \\

\textbf{Competing interests:} 
The authors declare no association in the development of AI models by OpenAI, Google, Anthropic or xAI. However, we confess to being daily users of these systems for email drafting, slide polishing, and other non-clinical generalist purposes and sometimes even to debug our LaTeX.\\

\textbf{Data availability:} 
The \textit{Radiology's Last Exam (RadLE) v1} dataset used in this study is not publicly released to prevent potential future model contamination. However, we will continue to evaluate emerging frontier models and release updated results as they become available. Research groups or medical AI labs wishing to benchmark their own models on this dataset may contact the corresponding authors. Upon request, we can provide a submission schema and evaluation specification so that groups can run their models and return model output. The ground truths remain withheld and scoring will be performed internally by radiologists from CRASH Lab to ensure fairness and data integrity. In the future, we plan to expand this into a larger global dataset in which case a part of the RadLE dataset and reasoning traces will be made public. We welcome interested research collaborators who want to participate in the larger study.\\

For testing models on the dataset, the corresponding authors can be reached at:  
\texttt{suvrankar.datta@ashoka.edu.in},  
\texttt{divyabuchireddy@gmail.com}, and  
\texttt{crashlab.kcdha@gmail.com}. \\

\textbf{Annexures:} 
Annexure~1 shows examples of reasoning traces on two representative cases across all five frontier AI models.
\\

\bibliography{references}  

\newpage
\section*{Annexure}
\addcontentsline{toc}{section}{Annexure} 



\section*{Supplementary material}

# Annexure

## Gemini 2.5 Pro

*User Prompt (Instructions)*          *Medical Image Input*

*You are a board-certified diagnostic radiologist. Given a medical image, your task is to provide the single most specific final diagnosis.*

*Instructions:*
*…*
*…*

*Example outputs (for demonstration only):*
*…*
*…*

*Please review the image and respond as instructed, with only the diagnosis in a single line.*

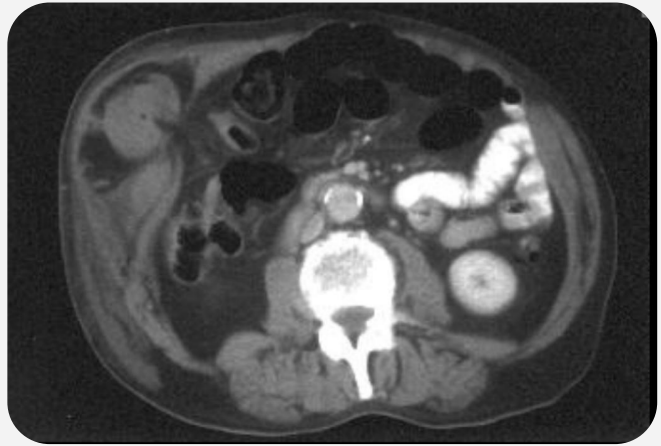

### Reasoning Trace

**Interpreting Abdominal CT**
I'm focusing on the **axial contrast-enhanced CT scan of the abdomen**. My initial assessment centers on identifying any key areas of contrast enhancement or abnormalities in the abdominal organs. Currently, I'm examining the liver, spleen, pancreas, kidneys, and major vessels for any focal lesions or signs of inflammation.

**Diagnosing the Lesion**
I've examined the CT scan closely. The **well-defined, oval-shaped lesion in the right lower quadrant, with its characteristic "target" or "whorled" appearance of alternating fat and soft tissue**, immediately caught my attention. Located in the greater omentum and accompanied by surrounding inflammatory stranding, it strongly suggests a diagnosis of **omental infarction**. While other possibilities exist, the image findings strongly support this, given its presentation, location, and the visualized tissue densities.

**Revisiting Differential Diagnoses**
I've refined the differential diagnosis based on the imaging findings. While omental infarction remains the most likely explanation for this characteristic lesion with its whorled appearance and inflammatory changes in the greater omentum, I'm considering other possibilities with more precision. **Epiploic appendagitis** is less likely given the lesion's size and complexity. However, I must exclude rarer entities such as a **fat-containing tumor** or **resolving hematoma**, though they are less probable given the typical omental infarction presentation.

**Concluding with Certainty**
Based on the distinctive findings, I'm confident in the diagnosis. The lesion's location, size, and particularly its "target" appearance with surrounding inflammatory changes, strongly support **omental infarction**. Ruling out other possibilities, I've concluded with certainty, that this is indeed an omental infarction.

### Diagnosis

**Omental infarction** 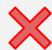

*Representative Output from different AI models on Case RA10 (Spigelian Hernia)*

## OpenAI o3

*User Prompt (Instructions)*  *Medical Image Input*

*You are a board-certified diagnostic radiologist. Given a medical image, your task is to provide the single most specific final diagnosis.*

*Instructions:*
*…*
*…*

*Example outputs (for demonstration only):*
*…*
*…*

*Please review the image and respond as instructed, with only the diagnosis in a single line.*

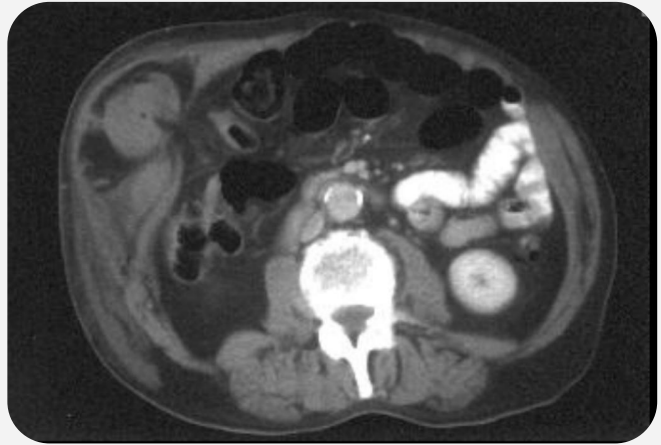

---

### *Reasoning Trace*

I'm analyzing a **CT of the abdomen.** There's a **circular lesion near the right side** — **but is that the patient's left due to image orientation?** It could be a **renal cyst,** though there also seems to be a **"target sign" in the mesentery**, possibly indicating **intussusception**. And on the right side, is that a **"whirlpool" or "target" sign in the mesenteric fat**? I'll zoom in for a closer look.

I'm noticing **multiple small bowel loops with wall enhancement, and some have a target shape.** On the right, there's a **round ring of hyperdensity with central fat** — perhaps indicative of intussusception. I'll zoom in for a closer look to get more details and ensure accuracy.

I'm seeing a **"target sign" in the right flank**, which could point to small bowel intussusception. Specifically, it might be **jejunojejunal intussusception.** Based on this CT slice showing the mid abdomen and kidney visibility, I'll go with "intestinal intussusception." It satisfies the request for a specific yet clear diagnosis.

I'm seeing a **ring structure with what looks like mesenteric fat in the center, and multiple rings elsewhere**. Intussusception comes to mind, possibly highlighted on a barium meal. The **left kidney shows a large, low-density lesion, which could be a simple cyst,** though renal cell carcinoma is more likely on a diagnostic quiz. I'll need to focus on the small bowel for now, as that seems most relevant. I'll zoom in on the left kidney area to get a clearer view of the lesion and assess its characteristics more closely. This will help me confirm whether it's a simple cyst or if there's something more concerning to consider. Let's focus on getting a detailed look.

It looks like the kidney region, but I'm not entirely sure yet. To confirm, I'll zoom out and make the image bigger to get a broader view. That should give me a better sense of what's going on and help clarify whether this is really a kidney lesion.

The **area near the kidney seems to show low attenuation, which could be related to hydration**—but the CT might be non-contrast, making it difficult to interpret. The **'target' ring sign is** more likely to indicate intussusception. However, another possbility could be closed loop obstruction. I'll zoom further to clarify.

---

### *Diagnosis*

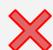

**Intussusception**

---

*Representative Output from different AI models on Case RA10 (Spigelian Hernia)*

# GPT-5

*You are a board-certified diagnostic radiologist. Given a medical image, your task is to provide the single most specific final diagnosis.*

*Instructions:*
*…*
*…*

*Example outputs (for demonstration only):*
*…*
*…*

*Please review the image and respond as instructed, with only the diagnosis in a single line.*

**Medical Image Input**

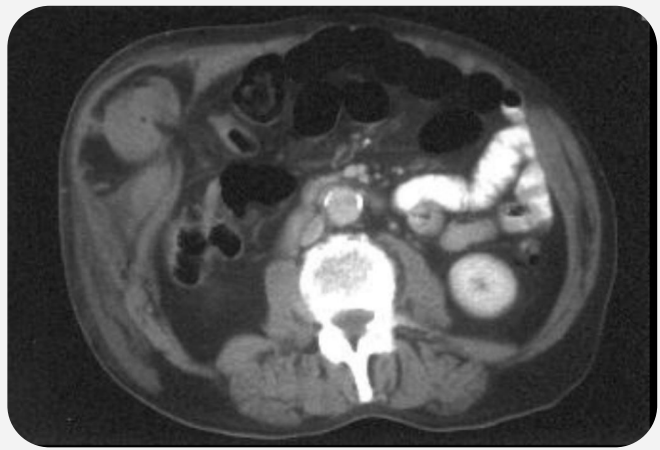

### Reasoning Trace

I'm looking at the **CT abdominal image**. **There's a circular shape and swirling of mesenteric vessels, which makes me wonder if it's a "whirlpool sign" indicating midgut volvulus.** Could be small bowel volvulus, but the patient already had ileocolic intussusception, so I'm considering it in terms of mesenteric malrotation too. Let's closely analyze this.

I'm seeing a round, **"target-like" loop on the right side, along with possible inversion of the SMA and SMV orientation**—though that's not clear in a single slice. While intussusception is a possibility, they've had one before. The "target sign" with concentric rings and the swirling mesenteric vessels make me think of a whirl sign, possibly indicating small bowel involvement.

The **"small bowel feces sign" suggests mechanical small bowel obstruction, with mixed gas and particulate matter in dilated loops.** The image shows "feculent" content, and could indicate this sign. However, given previous diagnoses like Crohn's, I'm leaning toward **midgut volvulus or SMA occlusion**. Omental torsion is another possibility.

### Diagnosis

**Midgut volvulus** ❌

*Representative Output from different AI models on Case RA10 (Spigelian Hernia)*

## Claude Opus 4.1

### User Prompt (Instructions)

*You are a board-certified diagnostic radiologist. Given a medical image, your task is to provide the single most specific final diagnosis.*

*Instructions:*
*…*
*…*

*Example outputs (for demonstration only):*
*…*
*…*

*Please review the image and respond as instructed, with only the diagnosis in a single line.*

### Medical Image Input

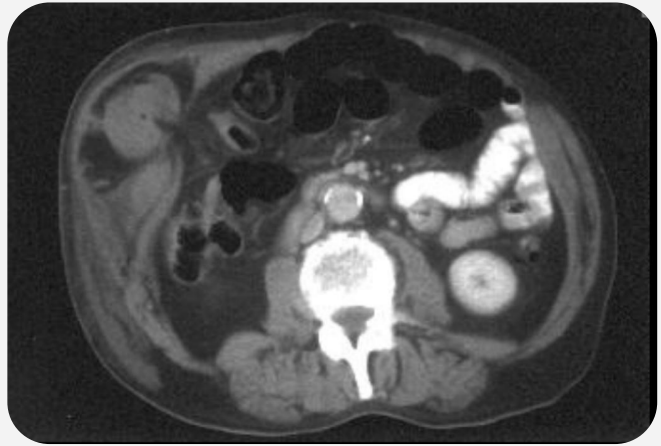

### Reasoning Trace

This is an **axial CT image of the abdomen**. I can see **multiple fluid-filled structures in the bowel that appear to be dilated small bowel loops.** There's significant bowel dilatation visible with multiple air-fluid levels. The pattern suggests small bowel obstruction.

Looking more carefully, I can see what appears to be **a transition point in the left side of the abdomen where the bowel caliber changes from dilated to collapsed.** This is characteristic of mechanical small bowel obstruction.

The most specific diagnosis I can provide based on this single image would be **small bowel obstruction**. However, I cannot determine the specific cause (adhesions, hernia, tumor, etc.) from this image alone.

### Diagnosis

**Small bowel obstruction** ❌

*Representative Output from different AI models on Case RA10 (Spigelian Hernia)*

# Grok 4

*You are a board-certified diagnostic radiologist. Given a medical image, your task is to provide the single most specific final diagnosis.*

*Instructions:*
*…*
*…*

*Example outputs (for demonstration only):*
*…*
*…*

*Please review the image and respond as instructed, with only the diagnosis in a single line.*

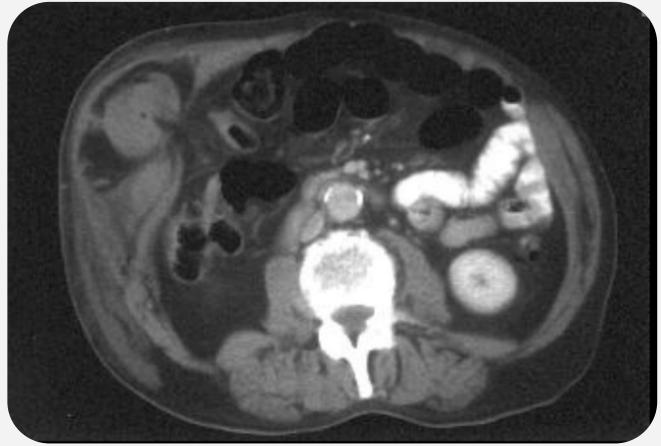

---

## *Reasoning Trace*

**Analyzing medical image**

- The task is to provide the most specific diagnosis from a medical image.
- It's a **CT scan of the abdomen, likely with contrast**, showing the spine and organs.
- The **vertebral body appears in the center bottom**, bright white bone.
- Spleen is on the right of the image, smaller than other organs.
- The image can be analyzed directly since it's uploaded, no URL needed.

**Examining abdominal structures**

- The aorta is small and round, bright in front of the spine.
- **Oral contrast is visible in the small bowel,** with the **duodenum opacified.**
- The **aorta appears enlarged, about 4–5 cm**, with a **calcified wall.**
- Let's check if it's a non-contrast CT, as the **aorta center isn't bright.**
- **The liver and spleen are visible, suggesting the upper abdomen at T12–L1 level.**

**Evaluating aortic features**

- The aorta's center looks **brighter, possibly indicating contrast enhancement.**
- The structure in front of the spine is the aorta, enlarged, with central bright contrast-enhanced lumen.
- The main abnormality is the **enlarged aorta with possible thrombus**.
- The aorta shows a bright lumen, **dark thrombus around it**, and a thin bright calcified wall.

---

## *Diagnosis*

**Abdominal aortic aneurysm** ❌

---

*Representative Output from different AI models on Case RA10 (Spigelian Hernia)*

## Gemini 2.5 Pro

**User Prompt (Instructions)**

*You are a board-certified diagnostic radiologist. Given a medical image, your task is to provide the single most specific final diagnosis.*

*Instructions:*
*…*
*…*

*Example outputs (for demonstration only):*
*…*
*…*

*Please review the image and respond as instructed, with only the diagnosis in a single line.*

**Medical Image Input**

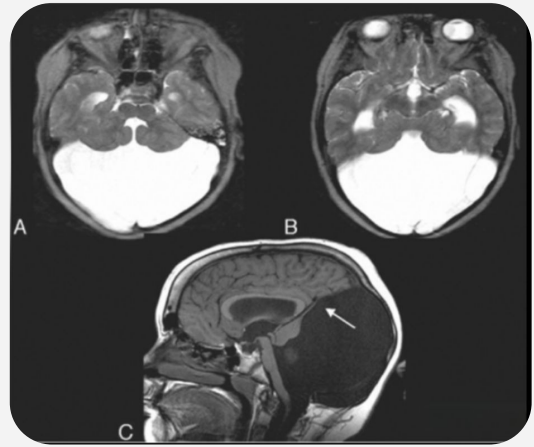

---

### Reasoning Trace

**Considering Brain MRI Data**
I'm focusing intently on analyzing the **three MRI scans of the brain.** I'm carefully reviewing each section to discern any potential anomalies or indications of pathology. My diagnostic process involves a methodical examination of the images, evaluating tissue contrast, and identifying areas of concern.

**Evaluating Current Observations**
My attention is now on the specific findings gleaned from the MRI scans. I'm focusing on the **large, CSF-filled space in the posterior fossa.** The cerebellar hemispheres' small size and the **elevated tentorium cerebelli** are also important. The **enlarged ventricles further confirm the hydrocephalus**, as expected. The fourth ventricle's communication with the cyst is another significant observation in this case.

**Diagnosing Dandy-Walker**
My evaluation continues, focusing on the distinct radiological features. The **significant vermian hypoplasia, extensive posterior fossa cyst communicating with the enlarged fourth ventricle, and the associated hydrocephalus** are all strong indicators. The image characteristics align perfectly with the classic triad of Dandy-Walker malformation. Mega cisterna magna and Blake's pouch cyst are ruled out due to the vermian findings. Arachnoid cyst isn't as fitting due to the distinct communication.

**Finalizing the Diagnosis**
My evaluation process is now complete. I've considered all pertinent radiological findings, and synthesized them to rule out alternative diagnoses. The communication between the cyst and the fourth ventricle, coupled with the vermian hypoplasia and hydrocephalus, aligns perfectly with the diagnostic criteria. Based on the conclusive imaging characteristics, I confidently conclude that the diagnosis is **Dandy-Walker malformation**.

---

### Diagnosis

**Dandy-Walker malformation** ✅

---

*Representative Output from different AI models on Case RA2 (Dandy Walker Malformation)*

## OpenAI o3

**User Prompt (Instructions)**

*You are a board-certified diagnostic radiologist. Given a medical image, your task is to provide the single most specific final diagnosis.*

*Instructions:*
*…*
*…*

*Example outputs (for demonstration only):*
*…*
*…*

*Please review the image and respond as instructed, with only the diagnosis in a single line.*

**Medical Image Input**

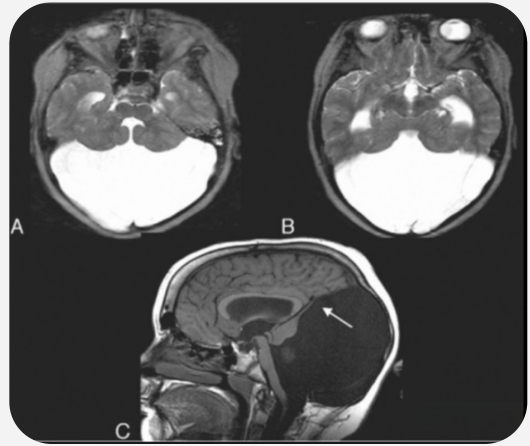

### Reasoning Trace

The **T1-weighted midline sagittal MRI** shows a **large cystic space replacing the cerebellum and vermis, with possible communication with the fourth ventricle.** This could indicate Dandy-Walker malformation, with a **typical absent vermis and upward displacement**. An arachnoid cyst or post-endarachnoid meningocele is also possible, but the features leans towards Dandy-Walker.

In Dandy-Walker malformation, the cyst is a dilation of the fourth ventricle, with cerebellum and vermis hypoplastic and a raised tentorium, resulting in an enlarged posterior fossa. An arachnoid cyst wouldn't show vermian hypoplasia. Given the image's **cystic area and brainstem positioning**, **Dandy-Walker malformation** is the most likely diagnosis.

### Diagnosis

**Dandy-Walker malformation** ✅

*Representative Output from different AI models on Case RA2 (Dandy Walker Malformation)*

# GPT-5

*You are a board-certified diagnostic radiologist. Given a medical image, your task is to provide the single most specific final diagnosis.*

*Instructions:*
*…*
*…*

*Example outputs (for demonstration only):*
*…*
*…*

*Please review the image and respond as instructed, with only the diagnosis in a single line.*

*Medical Image Input*

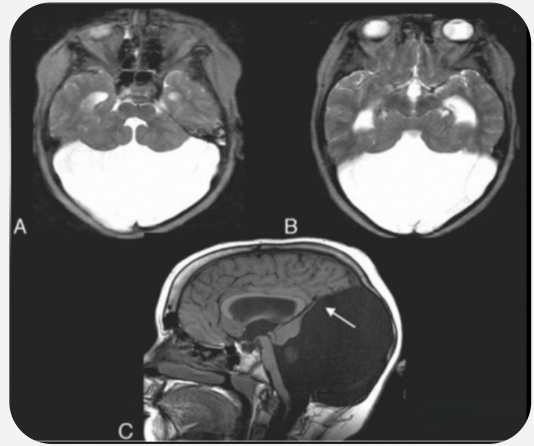

## Reasoning Trace

The image includes **axial T2 and mid-sagittal T1 scans**. In the T1 image, there's a large cystic lesion in the posterior fossa, possibly in the occipital region. The **arrow points to a mural nodule**. This could suggest a **cerebellar pilocytic astrocytoma** (in children) or **hemangioblastoma** (more common in adults). Let's inspect further.

The image shows a **hyperintense cyst in the posterior fossa**. If it's in an adult, hemangioblastoma is a possibility, especially with a mural nodule. But it could be an arachnoid cyst, though those don't have mural nodules. There's also a possibility of a Dandy-Walker malformation or "mega cisterna magna." Let's dig deeper.

The lesion could be a **posterior fossa arachnoid cyst, extending extra-axially and causing mass effect on the cerebellum.** The axial images show a well-defined CSF lesion, but I can't confirm FLAIR suppression. The sagittal T1 boundary near the occipital bone and linear structure could indicate an epidermoid cyst, though those typically show different properties on FLAIR and diffusion.

Panel C shows a large extra-axial cystic mass near the midline of the posterior fossa. There's a **small calcified meniscus and possible dural defect near the occipital bone.** Considering its appearance, this could be a posterior fossa arachnoid cyst—it's CSF signal on both T1 and T2 with no mural nodule, fitting the pattern for this diagnosis.

Given the mass effect on the cerebellum and displacement, the lesion is more likely a posterior fossa arachnoid cyst. A mega cisterna magna wouldn't cause this, and the image doesn't fit Dandy-Walker or other conditions. So, the most specific diagnosis is a **"posterior fossa arachnoid cyst."**

## Diagnosis

**Posterior fossa arachnoid cyst** ❌

*Representative Output from different AI models on Case RA2 (Dandy Walker Malformation)*

## Claude Opus 4.1

**User Prompt (Instructions)**

*You are a board-certified diagnostic radiologist. Given a medical image, your task is to provide the single most specific final diagnosis.*

*Instructions:*
*…*
*…*

*Example outputs (for demonstration only):*
*…*
*…*

*Please review the image and respond as instructed, with only the diagnosis in a single line.*

**Medical Image Input**

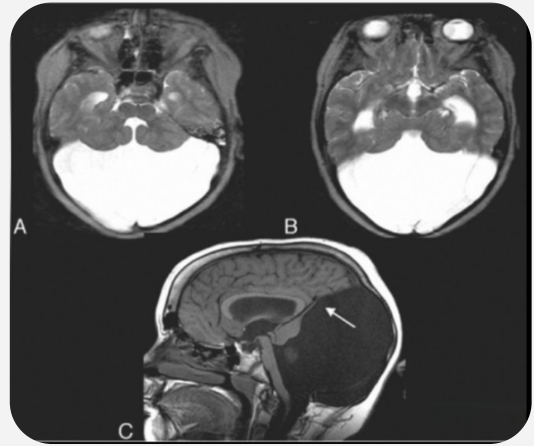

### Reasoning Trace

Looking at these MRI images:
Image A and B appear to be **axial T2-weighted MRI images of the brain**, showing the posterior fossa/cerebellum level.
Image C is a **sagittal T1-weighted MRI image of the brain** with an arrow pointing to what appears to be an abnormality.
In images A and B (axial views), I can see:
The cerebellum appears to have abnormal signal
There seems to be **increased T2 signal in the cerebellar hemispheres bilaterally**
The **fourth ventricle appears prominent**
In image C (sagittal view), the arrow is pointing to what appears to be:
An **abnormality in the region of the corpus callosum**
It looks like there might be **absence or thinning of the corpus callosum**
The combination of:
Cerebellar abnormalities (appears to be cerebellar atrophy or hypoplasia)
Corpus callosum abnormality (agenesis or dysgenesis)
The pattern on T2 imaging
This constellation of findings is most consistent with Dandy-Walker malformation or variant. The key features are:
Cerebellar vermis hypoplasia/absence
Cystic dilatation of the fourth ventricle
Enlarged posterior fossa
Actually, looking more carefully, the sagittal image shows what appears to be a **large posterior fossa cyst communicating with the fourth ventricle, and the cerebellar vermis appears hypoplastic or absent.** The arrow is pointing to the region where we would expect to see normal cerebellar structures.
This is classic for **Dandy-Walker malformation.**

### Diagnosis

**Dandy-Walker malformation** ✅

*Representative Output from different AI models on Case RA2 (Dandy Walker Malformation)*

# Grok 4

**User Prompt (Instructions)**

*You are a board-certified diagnostic radiologist. Given a medical image, your task is to provide the single most specific final diagnosis.*

*Instructions:*
*…*
*…*

*Example outputs (for demonstration only):*
*…*
*…*

*Please review the image and respond as instructed, with only the diagnosis in a single line.*

**Medical Image Input**

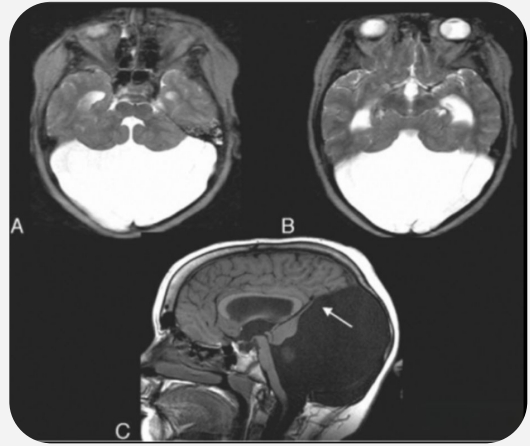

---

**_Reasoning Trace_**

**Analyzing request**

The task is to act as a board-certified diagnostic radiologist and provide a specific diagnosis from a medical image.

Output must be the single most specific diagnosis in full words, with no abbreviations or extra text.

The image includes **three MRI scans: A and B are axial brain views, and C is a sagittal view with an arrow in the posterior fossa.**

A and B show axial views through the posterior fossa, with A likely T1-weighted and B T2-weighted.

A shows a **hypointense fluid-filled space in the posterior cerebellum, enlarged fourth ventricle, and hypoplastic vermis.**

**Evaluating diagnosis**

The images suggest **Dandy-Walker malformation,** with a large cyst and small vermis shown in C.

Arachnoid cyst is a differential, but it doesn't typically cause vermis hypoplasia.

---

**_Diagnosis_**

**Dandy-Walker malformation** ✅

---

*Representative Output from different AI models on Case RA2 (Dandy Walker Malformation)*



\end{document}